\documentclass[11pt]{csdc}
\usepackage{algorithm}
\usepackage{algorithmicx}
\usepackage[noend]{algpseudocode}
\usepackage{microtype}
\usepackage{graphicx}
\usepackage{amssymb}

\makeatletter
\def\BState{\State\hskip-\ALG@thistlm}
\makeatother

\algnewcommand{\LineComment}[1]{\State$\triangleright$ #1}

\title{Anomaly detection and motif discovery in symbolic representations of time series}
\author{Fabio Guigou \and Pierre Collet \and Pierre Parrend}

\date{\today}
\newcommand{\subtitle}{TECHNICAL REPORT N\textsuperscript{o}69427/2}
\newcommand{\organisation}{4PFactory: E-Laboratory on the Factory of the Future}
\newcommand{\institutionalreportnumbers}{ARK unitwin-cs.org/ark:/69427/02}
\newcommand{\publisher}{Publisher \\
  Complex System Digital Campus \\
  UNITWIN UNESCO \\
  \urlstyle{same}\url{http://unitwin-cs.org} \\ \\
  \institutionalreportnumbers}

\begin{document}

\coverpage
\maketitle

\thispagestyle{empty}
\resume{Abstract}{
The advent of the Big Data hype and the consistent recollection of event logs and real-time data from sensors, monitoring software and machine configuration has generated a huge amount of time-varying data in about every sector of the industry. Rule-based processing of such data has ceased to be relevant in many scenarios where anomaly detection and pattern mining have to be entirely accomplished by the machine. Since the early 2000s, the de-facto standard for representing time series has been the Symbolic Aggregate approXimation (SAX). In this document, we present a few algorithms using this representation for anomaly detection and motif discovery, also known as pattern mining, in such data. We propose a benchmark of anomaly detection algorithms using data from Cloud monitoring software.
}
{Time series, Symbolic representation, Anomaly detection, Pattern mining}

\newpage\clearpage
  
\tableofcontents
\listoffigures
\listoftables

\chapter{Introduction}

\section{Problem statement}

We consider the problem of detecting anomalies and establishing baseline behaviour in time series captured by network monitoring software. Since the inception of such software, failure detection has mostly been performed by polling a number of metrics, storing them for expert analysis and applying simple decision rules based on the last few data points of each series independently, e.g. signaling a failure when all points within a short analysis window stay above a given threshold. While this approach has worked well enough to keep data-centers running, it still fails to capture many important events, such as deviations from a normal baseline or early signs of failure. While algorithms exist to perform such tasks, using them for monitoring would require significant hardware upgrades. We study alternative methods to fill the gap between crude expert systems and costly anomaly detection and pattern mining algorithms.

\section{Change and anomaly detection in time series}

\subsection{Symbolic representations}

Anomaly detection in time series is a prominent task in data-mining. However, the size and number of such series makes it extremely demanding in terms of computational power. To overcome this issue, many alternative representations have been proposed for time series: Discrete Fourier Transform, Wavelet Transform \cite{lin2004iterative}, Singular Value Decomposition... SAX was proposed in 2003 \cite{lin2003symbolic} as a simple, compact, text-based representation that reduces dimensionality and allows the use of string processing algorithms to analyze time series \cite{lin2007experiencing}. It has seen applications in time series indexing \cite{camerra2010isax}, visualization \cite{gu2013itree} and various mining tasks \cite{hung2007combining,rakthanmanon2013fast,senin2013sax}. Applications have even been tried on objects that are only remotely connected to time series, such as motion detection \cite{junejo2014silhouette}.

\subsection{On raw data}

Other statistical approaches have been used on raw time series data. Change point detection, i.e. detecting when a model stops fitting the data and a new one must be derived, has been implemented in many ways: using Bayesian models \cite{adams2007bayesian} or sequential testing \cite{blazek2001novel,tartakovsky2004change} to perform online detection. These methods have proven computationally efficient and able to reliably detect structural changes in data stream. However, they do not offer the wide range of applications that come with symbolic representation, and especially string representation. In this report, we focus on the opportunities offered by this paradigm.

\section{Context and goals}

In the ever growing field of Cloud computing and Cloud networking, service providers face the challenge of operating a high number of devices, both physical and virtual, while maintaining a contractually defined service level, known as SLA (Service Level Agreement). This situation calls for a tight monitoring of all the infrastructure. Monitoring software typically polls devices using ICMP and SNMP, collects data such as response time, CPU load, memory usage, and compares these values to thresholds. In the event of a value crossing the alert threshold, some form of warning is sent to the administrators.

While such software ``does the job'', it is often imprecise, yields lots of false positive or raises alerts too late, when the device has already failed or a customer has already opened a case for performance degradation. These shortcomings indicate the need for more advanced techniques, such as behavioral anomaly detection, that can either replace or improve the current methods.

The point of this study is to determine the feasibility of real-time anomaly detection in computer network monitoring time series. The advantages of such approach are expected to be low computational costs and good accuracy. Since monitoring, even using only crude threshold methods, is already CPU-intensive, we turned to symbolic representations of time series to search for a method compatible with low-resource monitoring servers.

\chapter{The SAX representation}

SAX is a time series representation designed to vastly reduce the data dimensionality and redundancy by subsampling and quantization. Typical settings use a subsampling factor $n$ of 8 to 10 and only $\alpha=3$ or $4$ bins for quantization. It is widely accepted that the impact of this parameter is small. The basic algorithm operates on the whole time series by applying a $z$-normalization (setting the mean to 0 and standard deviation to 1), replacing each non-overlapping window of $n$ points by its average -- a process known as Piecewise Aggregate Approximation (PAA) -- and applying a Gaussian quantization of this PAA into $\alpha$ bins.

It is worth noting that, while the conversion of a time series into a character string is not a familiar operation, the subsample-and-quantize process applied is the basis of any lossy compression algorithm: JPEG for image, MPEG for video, A-law and $\mu$-law for audio... The vector quantization, i.e. the dispatching of values into bins, depends on the original assumptions on the underlying statistical distribution. SAX \cite{lin2003symbolic} assumes a normal distribution of the time series values at the scale of a SAX word, as shown in Algorithm \ref{alg:sax_encoding}.

\begin{algorithm}
\caption{Basic SAX encoding algorithm}
\label{alg:sax_encoding}
\begin{algorithmic}

\Procedure{SAXEncode}{}
\BState input: 
\State	$S = [s_1, s_2, ..., s_N]$ \Comment{real-valued time series}
\State	$\alpha = int$ \Comment{cardinality (typical values: 3 or 4)}
\State	$n = int$ \Comment{symbol size (typical values: 8 to 10)}
\BState output:
\State	$s = string$ \Comment{SAX representation}
\BState algo:
\LineComment{GQF is the Gaussian Quartile Function}
\State	$breakpoints = GQF(1/(\alpha-1), 2/(\alpha-1), ..., (\alpha-2)/(\alpha-1)$)
\LineComment{(for $\alpha=3$, $breakpoints = [-0.43, 0.43]$; for $\alpha=4$, $breakpoints = [-0.67, 0, 0.67]$)}
\State	$S = (S - mean(S)) / std(S)$
\State	$S_{aggr} = [sum(s_1, s_2, ..., s_n)/n, sum(s_{n+1}, s_{n+2}, ..., s_{2n})/n, ...]$
\State	$s = quantize(S_{aggr}, breakpoints)$
\EndProcedure

\end{algorithmic}
\end{algorithm}

However, though this approximation is able to capture the main characteristics of a time series, it is not always appropriate for algorithms designed to operate on fixed-size sliding windows. In order to analyze long running series of thousands of points, the SAX encoding can be applied to such sliding windows as shown in Algorithm \ref{alg:sax_sliding_window}.

\begin{algorithm}
\caption{The most widely used SAX encoding, using a sliding window}
\label{alg:sax_sliding_window}
\begin{algorithmic}

\Procedure{SAX}{}
\BState input:
\State	$S = [s_1, s_2, ..., s_N]$ \Comment{long real-valued time series}
\State	$\alpha = int$ \Comment{cardinality}
\State	$n = int$ \Comment{symbol size}
\State	$w = int$ \Comment{sliding window size (feature size, usually user-specified)}
\BState output:
\State	$s' = [s'_1, s'_2, ..., s'_{N-w}]$ \Comment{list of SAX words}
\BState algo:
\LineComment{sliding window extraction}
\State	$windows = [[s_1, s_2, ..., s_w], [s_2, s_3, ..., s_{w+1}], ..., [s_{N-w}, s_{N-w+1}, ..., s_N]]$
\State	$s' = SAXEncode(windows, \alpha, n)$
\EndProcedure

\end{algorithmic}
\end{algorithm}

All following algorithms use this representation (list of words with full overlap) unless otherwise stated. The sliding window is often referred to as a ``feature window'' in the literature because the window size is tuned to the approximate size of the feature one wishes to extract from the series. This is particularly true for periodic time series, such as ECG datasets or sound analysis \cite{kasetty2008real}.

\chapter{Anomaly detection}

\section{Hot SAX}

Hot SAX \cite{keogh2005hot} is not an anomaly detection algorithm \textit{per se}. Instead, it is a heuristic based on the SAX representation of a time series to accelerate the brute-force algorithm. The ``obvious'' way to detect an anomaly is to search for the subsequence with the highest distance to any other subsequence, i.e. finding $argmax_i(min_j(dist(x_i, x_j)), |i-j| > n$ over the set $x$ of all sliding windows of $n$ points extracted from the time series $t$. This search involves a quadratic number of distance computations and therefore is not suitable for more than a few hundred points. Note that self-matches are excluded: the nearest neighbor of a subsequence is only searched among other subsequences having no point in common with it. The naive implementation is shown in Algorithm \ref{alg:brute_force_anomaly_detection}:

\begin{algorithm}
\caption{Brute-force anomaly detection}
\label{alg:brute_force_anomaly_detection}
\begin{algorithmic}

\Procedure{Brute force discovery}{}
\BState input:
\State	$T = [t_1, t_2, ..., t_N]$ \Comment{time series}
\State	$n = int$ \Comment{window size}
\BState output:
\State	$dist = float$ \Comment{max Euclidean distance between any 2 subsequences}
\State	$loc = int$ \Comment{location of most anomalous subsequence}
\BState algo:
\State	$dist = 0$
\State	$loc = null$
	\For{$p = 1$ to $N$}[0]
		\State $nndist = inf$
		\For{$q = 1$ to $N - n + 1$}
			\If{$|p - q| \le n$}
				\State next
			\EndIf
			\If{$dist([t_p, t_{p+1}, ..., t_{p+n-1}], [t_q, t_{q+1}, ..., t_{q+n-1}]) < nndist$}
				\State $nndist = dist([t_p, t_{p+1}, ..., t_{p+n-1}], [t_q, t_{q+1}, ..., t_{q+n-1}])$
			\EndIf
		\EndFor
		\If{$nndist > dist$}
			\State $dist = nndist$
			\State $loc = p$
		\EndIf
	\EndFor
\EndProcedure

\end{algorithmic}
\end{algorithm}

However, changing the order in which distances are computed, while not changing the end result, may speed up the search by multiple orders of magnitude (while retaining a complexity of $O(N^2)$) by allowing early terminations of the loops, as shown in Algorithm \ref{alg:anomaly_detection_heuristics}.

\begin{algorithm}
\caption{Anomaly detection, with heuristics}
\label{alg:anomaly_detection_heuristics}
\begin{algorithmic}

\Procedure{Hot SAX}{}
\BState input:
\State	$T = [t_1, t_2, ..., t_N]$ \Comment{time series}
\State	$n = int$ \Comment{window size}
\BState output:
\State	$dist = float$ \Comment{max Euclidean distance between any 2 subsequences}
\State	$loc = int$ \Comment{location of most anomalous subsequence}
\BState algo:
\State	$dist = 0$
\State	$loc = null$
	\For{$p = 1$ to $N$ order by OuterHeuristic}
		\State $nndist = inf$
		\For{$q = 1$ to $N - n + 1$ order by InnerHeuristic}
			\If{$|p - q| \le n$}
				\State next
			\EndIf
			\If{$dist([t_p, t_{p+1}, ..., t_{p+n-1}], [t_q, t_{q+1}, ..., t_{q+n-1}]) < nndist$}
				\State $nndist = dist([t_p, t_{p+1}, ..., t_{p+n-1}], [t_q, t_{q+1}, ..., t_{q+n-1}])$
			\EndIf
			\If{$dist([t_p, t_{p+1}, ..., t_{p+n-1}], [t_q, t_{q+1}, ..., t_{q+n-1}]) < dist$}
				\State break
			\EndIf
		\EndFor
		\If{$nndist > dist$}
			\State $dist = nndist$
			\State $loc = p$
		\EndIf
	\EndFor
\EndProcedure

\end{algorithmic}
\end{algorithm}

Using the SAX representation, the outer and inner heuristics index the time series by its symbolic approximation and use the simple assumptions that rare SAX words correspond to rarely occurring subsequences, and that two similar sequences will have a similar SAX representation. Therefore, the outer heuristic first searches the most anomalous subsequence among thoses represented by the rarest SAX word, and the inner heuristic searches any subsequence's neighbors among other subsequences represented by the same word as shown in Algorithm \ref{alg:heuristics}.

\begin{algorithm}
\caption{Heuristics used for anomaly detection}
\label{alg:heuristics}
\begin{algorithmic}

\Procedure{OuterHeuristic}{}
\BState input:
	\State $T = [t_1, t_2, ..., t_N]$ \Comment{time series}
	\State $n = int$ \Comment{window size}
\BState output:
	\State $P = [p_1, p_2, ..., p_N]$ \Comment{ordered indexes}
	\State $H = \{word => [indexes]\}$ \Comment{indexed SAX representation of $T$}
\BState algo:
	\State $words = SAX(T, n)$
	\State $H = hash\{word => [$\textit{indexes of each occurrence of this word}$]\}$
	\State $min = min(hash.values)$ \Comment{(min = almost always 1)}
	\State $P = [$\textit{indexes of words occurring only min time}$] + [$\textit{other indexes, shuffled}$]$
\EndProcedure

\Procedure{InnerHeuristic}{}
\BState input:
	\State $T = [t_1, t_2, ..., t_N]$ \Comment{time series}
	\State $H = \{word => [indexes]\}$ \Comment{indexed SAX representation of $T$}
	\State $p = int$ \Comment{index of current subsequence}
\BState output:
	\State $Q = [q_1, q_2, ..., q_N]$ \Comment{ordered indexes}
\BState algo:
	\State $Q = H\{words[p]\} + [$\textit{other indexes, shuffled}$]$
\EndProcedure

\end{algorithmic}
\end{algorithm}

\section{Sequitur}

The Sequitur \cite{senin2015time} algorithm proposed by Nevill-Manning in 1997 \cite{nevill1997identifying} is a dictionary-based compression algorithm. It uses the concepts of symbol, rule and digram to build a compact representation of its input data. In Sequitur vocabulary, a symbol is either an input token (e.g. a single character, or byte) or a token representing a rule; a rule is a symbol standing for a digram; and a digram is a pair of symbols. Therefore, in a string ``abc'', ``a'', ``b'' and ``c'' are symbols and ``ab'' and ``bc'' are digrams. A rule will have the form ``A = ab'', meaning that a string ``Ac'' can be read as ``abc''. With such a rule, ``A'' and ``c'' are symbols and ``Ac'' is a digram that can be itself part of another rule.

Since Sequitur builds a compact, context-free generative grammar for any sequence, it has been used for various analytical tasks, such as program trace analysis \cite{walkinshaw2010using}, query of compressed XML databases \cite{lin2005supporting} or structure inference in DNA and musical pieces \cite{earl2003enhanced}. The algorithm has a complexity of $O(N)$, which makes it suitable even for large series.

Sequitur transforms any input sequence into a compact representation where two essentials constraints are met: no digram appears more than once in the output sequence (digram uniqueness), and no rule is used (either in the output sequence or in other rules) less than twice (rule utility). It is argued in \cite{senin2015time} that the number of rules used to represent a given point in a Sequitur-compressed time series, being proportional to its compressibility, is a good approximation of the series' Kolmogorov complexity, or algorithmic complexity (i.e. the size of the smallest program capable of generating the series) at that point. The assumption is that an anomaly is very likely to correspond to a rising complexity, or a lowering compressibility.

Sequitur builds a grammar tree in which the depth of a leaf is directly related to its frequency in the original string. Higher level, non-terminal nodes in the tree correspond to frequent patterns in the data. Conversely, shallow branches denote rare patterns, which is of interest in anomaly detection. With the analytic decompression routine that follows, the rule density, i.e. the depth in the grammar tree, can be associated to any point in the series from the Sequitur representation as shown in Algorithm \ref{alg:sequitur}.

\begin{algorithm}
\caption{Analytic unwrapping procedure}
\label{alg:sequitur}
\begin{verbatim}
unwrap(token, depth, rules) =
  if token not in rules then (token, depth)
  else [unwrap(rules{token}[1], depth + 1, rules),
    unwrap(rules{token}[2], depth + 1, rules)]

unwrap(blob, rules) = 
  unwrap(token, 0, rules) for each token in blob
\end{verbatim}
\end{algorithm}

The anomaly detection algorithm proposed uses as Sequitur symbols the SAX words representing subsequences (i.e. a basic symbol could be 'aabacd') as shown in Algorithm \ref{alg:sequitur_anomaly_detection}.

\begin{algorithm}
\caption{Anomaly detection using the Sequitur compression algorithm}
\label{alg:sequitur_anomaly_detection}
\begin{algorithmic}

\Procedure{SequiturAnomaly}{}
\BState input:
	\State $T = [t_1, t_2, ..., t_N]$ \Comment{time series}
	\State $n = int$ \Comment{window size}
\BState output:
	\LineComment{Sequitur rule density @ each point of the series (inverse anomaly score)}
	\State $density = [d_1, d_2, ..., d_N]$
\BState algo:
	\State $words = SAX(T, n)$
	\LineComment{SAX words [e.g. 'abbcab'] are used as elementary alphabet for Sequitur}
	\State $tokens, rules = Sequitur(words)$
		\State \textit{tokens = list of Sequitur symbols}
		\State \textit{rules = hash \{symbol $\rightarrow [symbol|letter, symbol|letter]$\}}
	\State $unwrapped = []$
	\ForAll{$token$ in $tokens$}
		\State $unwrapped << unwrap(token, 0, rules)$ \Comment{$unwrapped$ is [(SAX word, depth)]}
	\EndFor
	\For{$i = 1$ to $N$}
		\State $density = sum(unwrapped[max(i-n, 0), max(i-n+1, 0), ..., i])$
	\EndFor
\EndProcedure

\end{algorithmic}
\end{algorithm}

\section{Chaos game representation}

The Chaos Game representation \cite{wei2005assumption,barnsleyfractals} is a way of generating a bitmap from a DNA fragment, i.e. a string generated by an alphabet of four symbols (A, C, T, G). It recursively splits the two-dimensional space into pixels representing the number of occurrences of specific strings, adding one character at each level (e.g. at level one, it generates a 4-pixel image with the total count of A, C, G, and T bases. At level 2, the image is 4-pixel wide, with pixel corresponding to length-2 strings such as AA, AC, AT, ..., TG, TT). The proposed algorithm uses the SAX representation of a time series, with an alphabet size of 4, to build such bitmaps \cite{kumar2005time}, and then compare the bitmap of a detection window (after the currently analyzed point) and a lag window (before the point). Typically the lag window is 2 or 3 times longer than the detection window, therefore the bitmap must first be scaled.

Besides the graphical aspect of the bitmap generation, this algorithm is a simple histogram comparison: the frequency distribution of all possible $N$ symbols strings is computed to the left and right of each point in the time series and the two distributions are compared to yield an anomaly score. Since the number of bins in the histograms grows exponentially with the analysis level, the SAX sliding window will typically be short. The authors use a level 3 analysis, i.e. 64-bin histograms.

The anomaly score at any given point $t_i$ of the time series is given by $d_i = \sum_{j=1}^{N} (H_{i-lead, i} - H_{i, i+lag})^2$, where $H_{i, j}$ is the histogram computed between points $t_i$ and $t_j$, as shown in Algorithm \ref{alg:chaos_game}. Results presented in \cite{wei2005assumption} show that even subtle structural anomalies in periodic series can be detected.

\begin{algorithm}
\caption{Anomaly detection using the Chaos Game sequence representation}
\label{alg:chaos_game}
\begin{algorithmic}

\Procedure{ChaosGameAnomaly}{}
\BState input:
	\State $T = [t_1, t_2, ..., t_N]$ \Comment{time series}
	\State $n = int$ \Comment{symbol size}
	\State $w = int$ \Comment{window size}
	\State $D = int$ \Comment{detection window length (in number of SAX windows)}
	\State $L = int$ \Comment{lag window length (in number of SAX windows)}
\BState output:
	\State $score = [s_1, s_2, ..., s_N]$ \Comment{anomaly score}
\BState algo:
	\State words = SAX(T, 4, n, w)
	\For{$i = 1 + L$ to $N - D$}
		\State $det\_map = ChaosGame(words[i, i+n, ..., i+n*(D-1)]) / D$
		\State $lag\_map = ChaosGame(words[i-L*n, i-(L-1)*n, ..., i-n]) / L$
		\State $score_i = EuclidianDistance(det\_map, lag\_map)$
	\EndFor
\EndProcedure

\end{algorithmic}
\end{algorithm}

\chapter{Motif discovery}

\section{Minimal Description Length}

While not directly using the SAX representation, the Minimal Description Length algorithm \cite{vespier2013mining} transforms the time series in a symbolic representation by means of a multiscale Gaussian blur called scale-space image; at each level, segments bounded by the zero-crossings of the first derivative of the smoothed time series are extracted. These segments are compactly described by two coordinates: their length and the difference between their first and last point. This description is then quantized via a $k$-means clustering step. The time series at each scale can be processed as a string.

The Minimal Description Length (MDL) framework is an approach in which a motif is selected if it increases the compression ratio of the time series, i.e. it allows for a shorter description. Given the complexity of it exact computation (which would require the extraction and evaluation of all possible sets of motifs), a heuristic is used: only one motif is extracted at each scale. Typically, 8 scales are used. This algorithm, used on complex and noisy data sets such as vibration measured on a bridge, is able to isolate multiscale overlapping patterns, such as the passing of a single car on the aforementioned bridge as shown in Algorithm \ref{alg:min_descr_length}.

\begin{algorithm}
\caption{Motif detection using the Minimal Description Length framework}
\label{alg:min_descr_length}
\begin{algorithmic}

\Procedure{MDLMotif}{}
\BState input:
	\State $T = [t_1, t_2, ..., t_N]$ \Comment{time series}
	\State $n = int$ \Comment{decomposition level}
\BState output:
	\State $M = (m_1, m_2, ..., m_N)$ \Comment{set of motifs}
\BState algo:
	\State $M = ()$
	\For{$i = 1$ to $n$}
		\State $detail_i = T \star h_i$ \Comment{$h_i$ gaussian kernel of size $2^i$}
		\State $deriv_i = detail_{i,j+1} - detail_{i,j}, j \in \{1, 2, ..., N-1\}$ $1^{st}$ \Comment{derivative}
		\LineComment{Computation of zero-crossings of the derivative}
		\State $zC = \{j | deriv_{i, j} = 0\}$
		\State $segments_i = (\{zC_{k+1} - zC_k, detail_{i, zC_{k+1}} - detail_{i, zC_{k+1}}\}), k \in \{1, 2, ..., |zC| - 1\}$
		\State $motifs_i$ = set of repeating strings in $segments_i$
		\State $best\_motif_i = argmin_{m \in motifs_i} L(m) + L(T|m)$ \Comment{best predictor for $T$ at scale $i$}
		\State $M = M \cup best\_motif_i$
	\EndFor
\EndProcedure

\end{algorithmic}
\end{algorithm}

\section{Grammar inference}

As shown in the previous section, grammar inference \cite{li2010approximate} can be used not only to detect anomalies but, more naturally, to extract motifs. In this approach, the time series is SAX-encoded with a sliding window and converted into its Sequitur representation in the exact same way as for anomaly detection. The rules created during the compression are then mapped back to time series segments. Contrary to anomaly detection, the points with the highest rule density, i.e. the longest branches in the Sequitur grammar tree, are selected. A post-processing step is required to refine the results by:

\begin{itemize}
	\item eliminating self-matches (motif matching itself within a very small offset);
	\item selecting the longest patterns, which are more likely to convey useful information;
	\item removing ``obvious'' patterns, e.g. monotonically increasing or constant ones; and
	\item merging overlapping patterns if necessary
\end{itemize}

\section{MK algorithm}

The MK (Mueen-Keogh) \cite{mueen2009exact} algorithm is much related to HOTSAX, in that it uses a heuristic to prune an otherwise exact search for the most similar (instead of most different) pair of subsequences inside a time series. However, this algorithm is the only one in our collection operating on the raw time series data instead of a symbolic representation. We chose to include it in this report because of the similarity it bears to HOTSAX.

Instead of a SAX pre-processing to lower the number of distance evaluations, the algorithm uses reference points: subsequences chosen at random in the time series, and orders the other subsequences by their distance to the reference. The heuristic is based on the assumption that subsequences that are close to each other will also be close in this projection as shown in Algorithm \ref{alg:Mueen_Keogh}.

\begin{algorithm}
\caption{Mueen-Keogh algorithm}
\label{alg:Mueen_Keogh}
\begin{algorithmic}

\Procedure{MKMotif}{}
\BState input:
	\State $T = [t_1, t_2, ..., t_N]$ \Comment{time series}
	\State $n = int$ \Comment{motif length}
\BState output:
	\State $S_1, S_2$ \Comment{pair of subsequences with smallest distance}
\BState algo:
	\State $best\_dist = \infty$
	\State $ref = [T_r, T_{r+1}, ..., T_{r+n-1}]$, $r$ random
	\For{$i = 1$ to $N - n$}
		\State $dist_i = EuclidianDistance(ref, [T_i, T_{i+1}, ..., T_{i+n-1}])$
		\If{$dist_i < best\_dist$}
			\State $best\_dist = dist_i$
			\State $S_1 = ref$
			\State $S_2 = [T_i, T_{i+1}, ..., T_{i+n-1}]$
		\EndIf
	\EndFor
	\State Order subsequences $s_1, s_2, ..., s_{N-n}$ by their corresponding distances with function $I_j$ mapping to their original positions
	\State $\mathit{offset}=0$
	\State $abandon=0$
	\While{not abandon}
		\State $\mathit{offset} = \mathit{offset} + 1$
		\State $abandon=1$
		\For{$j=1$ to $N-n$}
			\If{$dist_{I_j} - dist_{I_{j+\mathit{offset}}} < best\_dist$}
				\State $abandon=0$
				\If{$EuclidianDistance(s_{I_j}, s_{I_{j+\mathit{offset}}}) < best\_dist$}
					\State $best\_dist = EuclidianDistance(s_{I_j}, s_{I_{j+\mathit{offset}}})$
					\State $S_1 = s_{I_j}$
					\State $S_2 = s_{I_{j+\mathit{offset}}}$
				\EndIf
			\EndIf
		\EndFor
	\EndWhile
\EndProcedure

\end{algorithmic}
\end{algorithm}

Note that the algorithm can use multiple reference points, using only one to perform the subsequence reordering and all of them to compute $best\_dist$. This further increases the convergence. As with HOTSAX, the overall complexity of the algorithm is not different from brute force search ($O(N^2)$), but the efficient pruning of distance computations makes the actual runtime orders of magnitude faster.

\section{Motif Tracking algorithm}

The Motif Tracking algorithm \cite{wilson2008motif}, shown in Algorithm \ref{alg:motif_tracking}, is an attempt to bring the Artificial Immune Systems (AIS) class of algorithms that operate almost only on strings into the field of time series processing. The SAX representation is used to produce the antibodies/antigens. It uses a subsequence length equal to the word size, therefore each subsequence is represented by a single symbol. Trackers are used as memory cells; they contain the string representation of a subsequence and a match count. Contrary to usual AIS algorithm, this one is deterministic.

\begin{algorithm}
\caption{The motif tracking algorithm}
\label{alg:motif_tracking}
\begin{algorithmic}

\Procedure{MotifTracking}{}
\BState input:
	\State $T = [t_1, t_2, ..., t_N]$ \Comment{first differential of time series}
	\State $\alpha = int$ \Comment{alphabet size}
	\State $s = 1$ \Comment{subsequence length}
	\State $r ) float$ \Comment{matching threshold (Euclidian distance)}
\BState output:
	\State $M = (m_1, m_2, ..., m_N)$ \Comment{set of motifs}
\BState algo:
	\LineComment{\textit{Trackers} are SAX alphabet symbols with associated scores and indexes of occurrences}
	\State $trackers = ([a, 0, {}], [b, 0, {}], [c, 0, {}], ...)$
	\State $symbols = [u_1, u_2, ..., u_{N-s}] = SAX(T, \alpha, s, s)$
	\State $trackers\_change = 0$
	\State $l = 1$ \Comment{motif length}
	\Repeat
		\For{$i = 1 to N-s-l$}
			\ForAll{$tracker, score, indexes \in trackers$}
				\If{$[u_i, u_{i+1}, ..., u_{i+l}] = tracker$}
					\If{$\forall j \in indexes, dist([t_i, t_{i+1}, ..., t_{i+s}], [t_{j}, t_{j+1}, ..., t_{j+s}]) / s < r$}
						\State $score += 1$
						\State $indexes = indexes \cup i$
					\EndIf
				\EndIf
			\EndFor
		\EndFor
		\ForAll{$tracker, score \in trackers$}
			\If{$score < 2$}
				\State Remove $tracker$ from $trackers$
			\Else
				\ForAll{$symbol \in alphabet$}
					\State Add $[tracker + symbol, 0]$ to $trackers$
				\EndFor
			\EndIf
		\EndFor
		\State $s += 1$
	\Until{$trackers\_change = 0$}
	\State $M = trackers$
\EndProcedure

\end{algorithmic}
\end{algorithm}

Since this algorithm detects exact motifs (without any mutation between any two occurrences) when they appear at least twice, it is logically equivalent to the grammar induction algorithm described earlier.

\section{Mining approximate motifs}

The `Mining approximate motifs' algorithm \cite{ferreira2006mining}, given in Algorithm \ref{alg:mining_approximate_motifs}, proposes a number of innovations: it uses aggregative clustering to detect motifs and Pearson's $r$ instead of Euclidean distance to match potential motif occurrences. Matches are encoded as 2-clusters that are later merged and extended. Note that the original algorithm is proposed for a database of time series; we show an adapted version for a single long time series.

\begin{algorithm}
\caption{Mining approximate motifs with aggregative clustering}
\label{alg:mining_approximate_motifs}
\begin{algorithmic}

\Procedure{AggregativeClusteringMotif}{}
\BState input:
	\State $T = [t_1, t_2, ..., t_N]$ \Comment{time series}
	\State $R_{min} = float$ \Comment{similarity threshold}
	\State $\alpha, word, window$ \Comment{SAX parameters}
\BState output:
	\State $M = (m_1, m_2, ..., m_N)$ \Comment{set of motifs}
\BState algo:
	\State $D = SAX(T, \alpha, word, window)$
	\State $clusters = \{\}$
	\For{$i = 1$ to $N$}
		\For{$j = i + window$ to $N - s$}
			\If{$|r(D_i, D_j)| \ge R_{min}$}
				\State $clusters = clusters \, \cup <i, j>$
			\EndIf
		\EndFor
	\EndFor
	\Repeat
		\State $updates = 0$
		\ForAll{$X = [x_1, x_2, ..., x_{|X|}] \in clusters$}
			\ForAll{$Y = [y_1, y_2, ..., y_{|Y|}] \in clusters \setminus X$}
				\If{$\forall x \in X \forall y \in Y |r(D_x, D_y)| \ge R_{min}$}
					\State $clusters = clusters \setminus X \setminus Y \, \cup <x_1, x_2, ..., y_1, y_2, ...>$
					\State $updates += 1$
				\EndIf
			\EndFor
		\EndFor
	\Until{$updates = 0$}
\EndProcedure

\end{algorithmic}
\end{algorithm}

The original algorithm proposes a way to extend the motif instances; however, our use of SAX sliding windows makes this step inconvenient and mostly useless: the motif length is an initial parameter. In this case, it is also possible to use SAX's lower-bound distance or Hamming distance as a similarity measure.

\chapter{Experimental evaluation}

\section{Input data}

To benchmark these algorithms, we use a collection of 14 time series spanning over two months, representing CPU load, memory usage, process count and active TCP sessions of 3 production servers and firewalls. The metrics are acquired with a temporal resolution of one point per minute. These series have been selected because they contain both identifiable motives and anomalies. In particular, they display various failure modes, some of which invalidate the basic assumptions of certain algorithms and therefore are not detected.

\begin{figure}
	\label{cpu2}
	\centering
	\includegraphics[width=15cm, height=4.5cm]{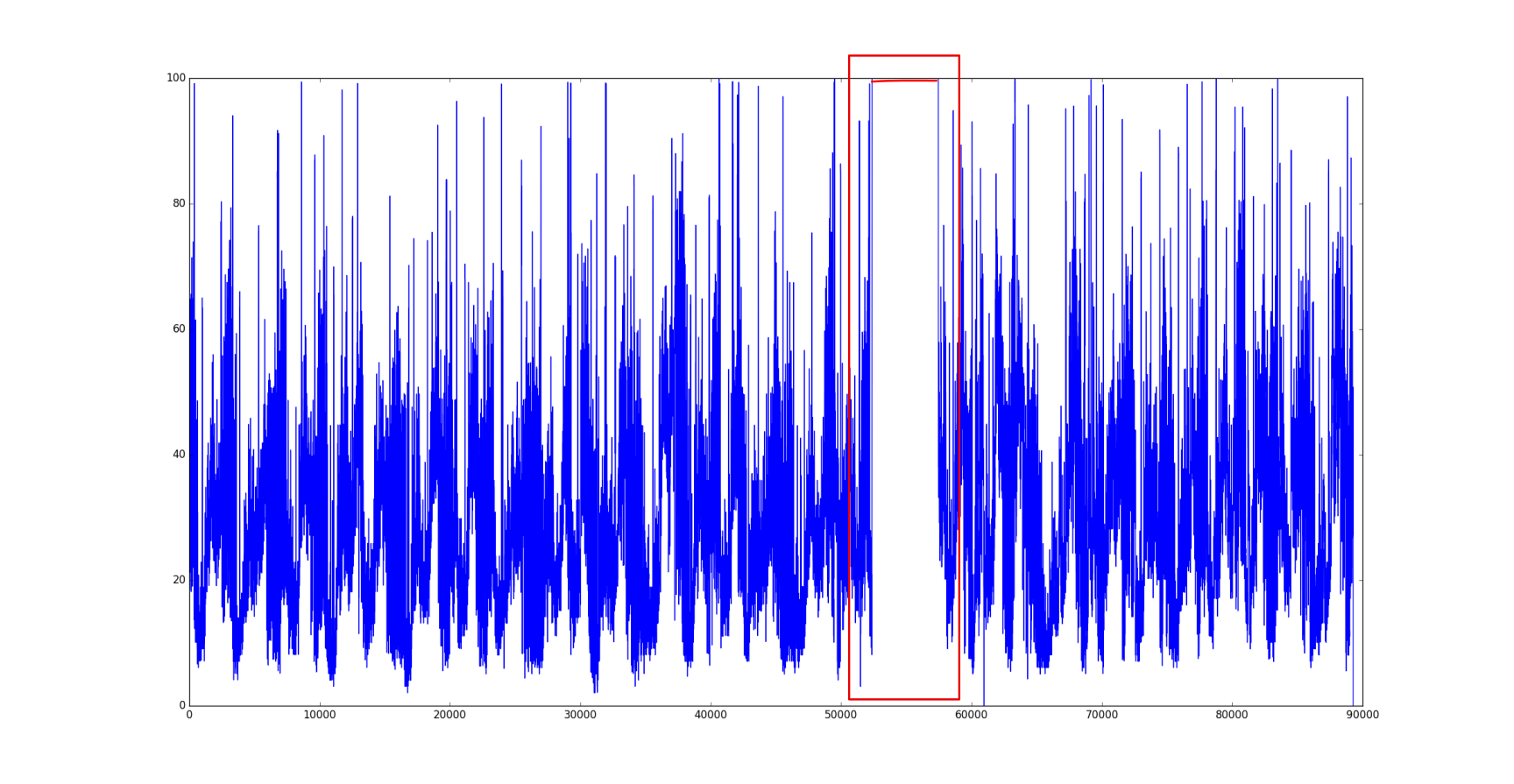}
	\caption{Anomaly in CPU load stuck at 100\%}
\end{figure}

\begin{figure}
	\label{fortiram}
	\centering
	\includegraphics[width=15cm, height=4.5cm]{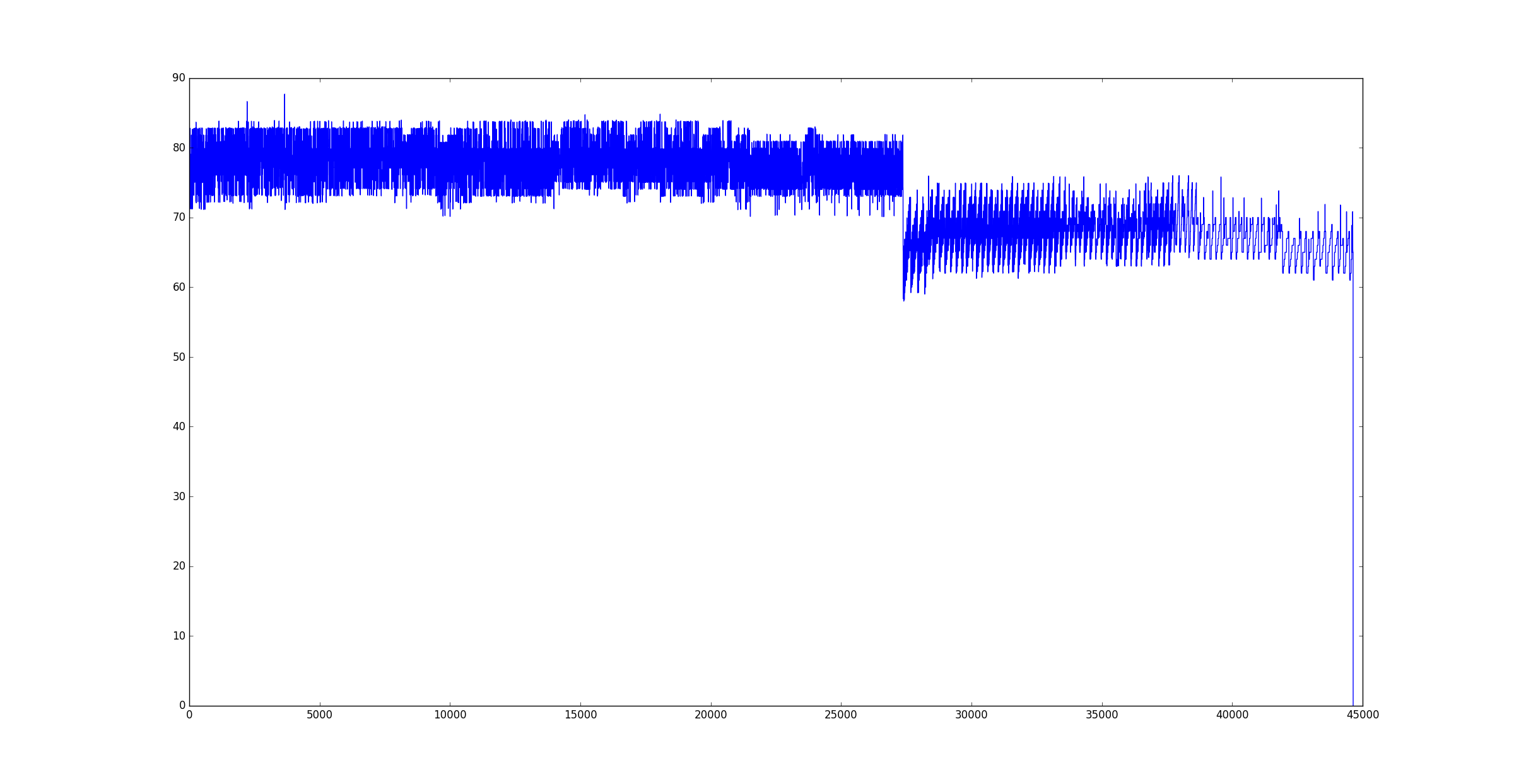}
	\caption{Transition in RAM usage (high and aperiodic to low and periodic)}
\end{figure}

Anomalies such as depicted in figure \ref{cpu2} are easy to detect even for conventional, commercial monitoring software. However, more subtle changes (figure \ref{fortiram}) that can be symptomatic of a critical component crashing are harder to detect and generally overlooked; instead, the crash might be detected later when it causes other symptoms and impact on other parts of the system. Detecting such types of anomalies is key to faster correction, smaller impact and easier root cause analysis.

\begin{figure}
	\label{memmotifs}
	\centering
	\includegraphics[width=15cm, height=4.5cm]{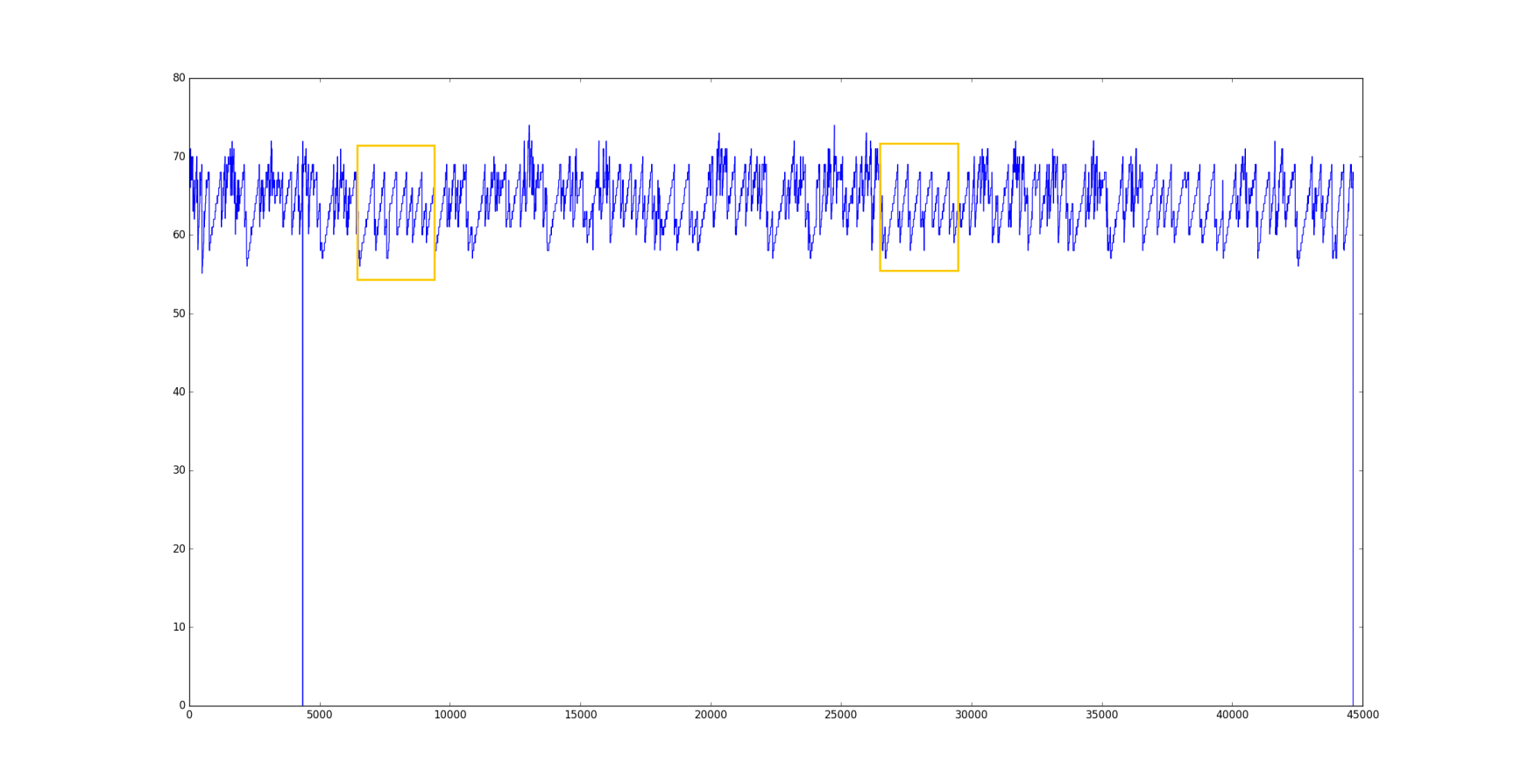}
	\caption{Small motif embedded in an apparently random RAM usage}
\end{figure}

\begin{figure}
	\label{ses}
	\centering
	\includegraphics[width=15cm, height=4.5cm]{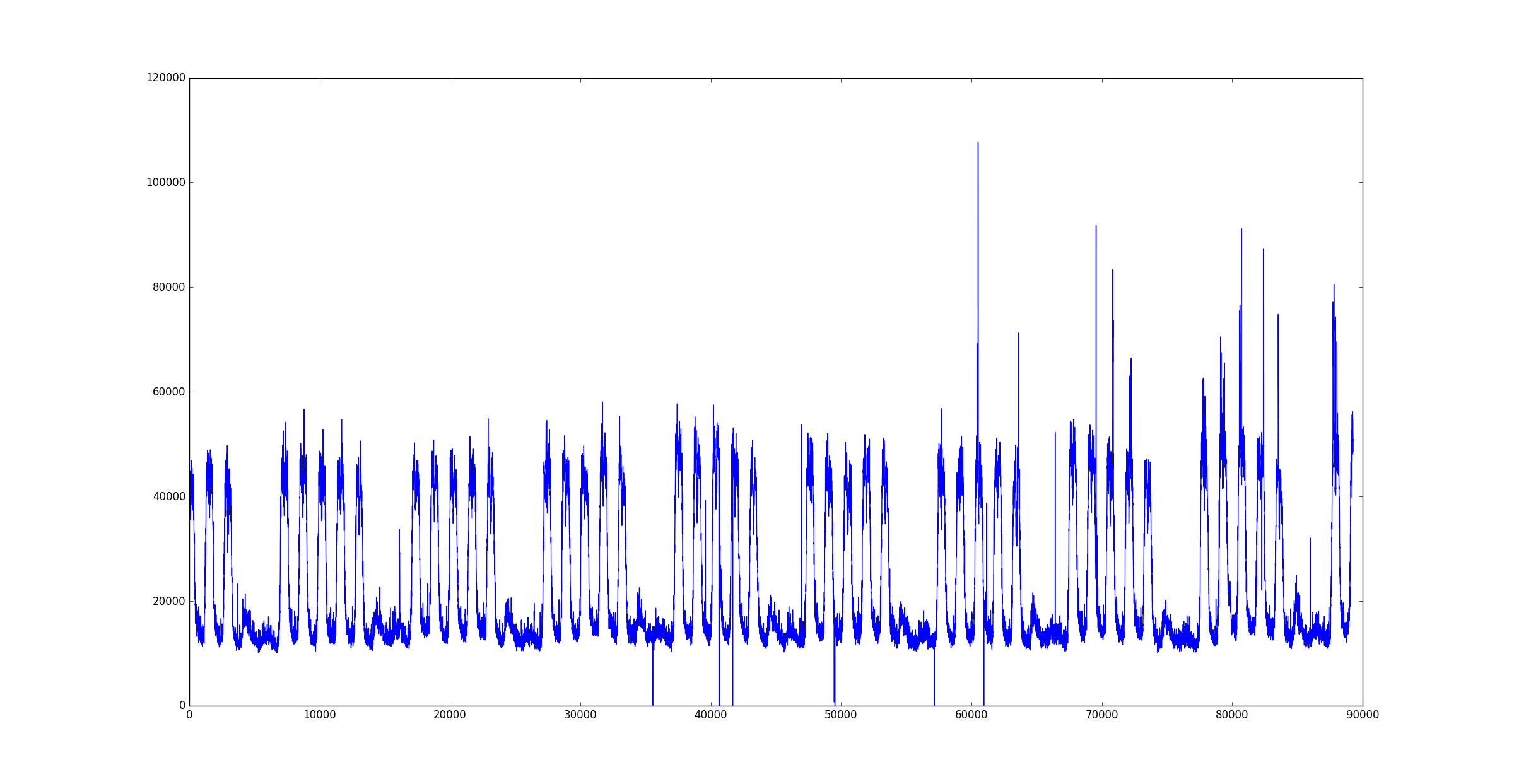}
	\caption{Obviously periodic TCP session count}
\end{figure}

In the context of IT network monitoring, awareness of motives can simplify analysis by two means: on the one hand, having an expert associate a label to a motif makes it possible to recognize a possibly complex and rare event in real time. On the other hand, periodicity makes traditional, threshold-crossing based methods unreliable because of the potentially large variation of the signal in its normal range. Being able to filter out large low-frequency components is a simple way to improve anomaly detection.

\section{Anomaly detection benchmark}

Since the very concept of an anomaly is hard to quantify (i.e. classifying data into ``normal'' or ``abnormal'' in time series can be somehow arbitrary, outside of the obvious transitions), we will perform a qualitative evaluation of these algorithms, taking into account execution speed (i.e. CPU load), failure modes (i.e. types of anomalies that cannot be detected), detection delay (i.e. number of data points required to spot an anomaly) and sensitivity to periodic input (i.e. precision loss due to the signal being periodic).

\subsection{Note on figure reading}

The blue graph is the raw time series, the green one the anomaly score and the red vertical lines indicate anomaly detection (anomaly score crossing a tunable threshold, here five standard deviations). The anomaly score does not necessarily begin at the same point as the series: it is delayed by buffering in the case of Chaos Game, which implies that the alarm is only raised after that buffering time.

\subsection{Performance figures}

We first study the running time of the algorithms. Since we expect to monitor thousands of time series on each monitoring server, a low running time is of paramount importance. Results show that Sequitur and Chaos Game yield about the same running time, between 6 and 7 seconds per month of data. The running time, as predicted, is linear with the size of the series. Hot SAX, on the other hand, displays a much larger, and quadratic, running time; we only ran it once on one month of data, and therefore have no estimate as to the variability of our result of almost 11 minutes. On a time series of two months, the algorithm ran for more than an hour without producing any result. Given that it definitely eliminates it as a viable candidate for IT monitoring, we did not try to let it complete.

\begin{table}
	\caption{Runtime comparison of anomaly detection algorithms (mean time (s) / standard deviation)}
	\centering
	\begin{tabular}{ | l | c | c | c | }
	\hline
	Nb of points & Chaos Game & Sequitur & Hot SAX\\ \hline
	44k & 6.66/0.24 & 6.62/3.03 & 651/?\\ \hline
	88k & 12.67/0.45 & 12.07/5.10 & ?/?\\ \hline
	\end{tabular}
\end{table}

\subsection{On SAX parameters}

As stated in almost all papers by Keogh et.al., the parameters used for SAX encoding (cardinality and word size) have little impact on the behaviour of any algorithm. At most, they can change the sensitivity and the running time, e.g. trade off accuracy for faster execution, or move along the precision-recall curve. Window size, however, must generally be adapted to the scale of any feature in the series, i.e. the size of the anomaly we try to detect, or the period of the signal.

This is one of the major shortcomings we observed in all SAX-based algorithms: with mostly weekly periodic signals, the best settings would delay any analysis until after at least 7 \textit{days} after a data point is acquired. This is due to the mismatch between the scale of the observed phenomena and the desired detection time. In most use cases, SAX is used to study phenomena of the scale of a second, with a few dozen points per occurrence. We study weekly patterns with thousands of points per occurrence; therefore, the periodic signal is mostly perceived as concept drift from the SAX perspective. However, even with this limitation, most anomalies can still be detected.

\subsection{Hot SAX}

As described above, Hot SAX is the only algorithm that does not generate an ``anomaly score'', or any kind of distance measure. It only returns the single most anomalous point in a series. While this point, in all our tests, always corresponded to the real anomaly (or \textit{one} of the anomalies), the algorithm remains extremely slow and its result is of little use in any real-time settings. In fact, our tests had trouble even completing, taking over 10 minutes for one single month of data (while the other algorithms only needed half a dozen seconds).

As Hot SAX, by its design, always outputs the mathematically defined worst anomaly, it has inherently no failure mode. In fact, it could be used in IT monitoring, with some restrictions:

\begin{itemize}
	\item The data must be undersampled.
	\item The search window must be limited.
	\item Hot SAX must only be applied to some critical metrics.
\end{itemize}

If the time series in which the search is conducted can be maintained within a few thousands of points, and Hot SAX is not used on each and every series, it can be a very precise tool for advanced analytics. Its CPU cost is the main obstacle to its adoption as a standard algorithm for anomaly detection.

\subsection{Sequitur}

\begin{figure}
	\label{seqok}
	\centering
	\includegraphics[width=15cm, height=4.5cm]{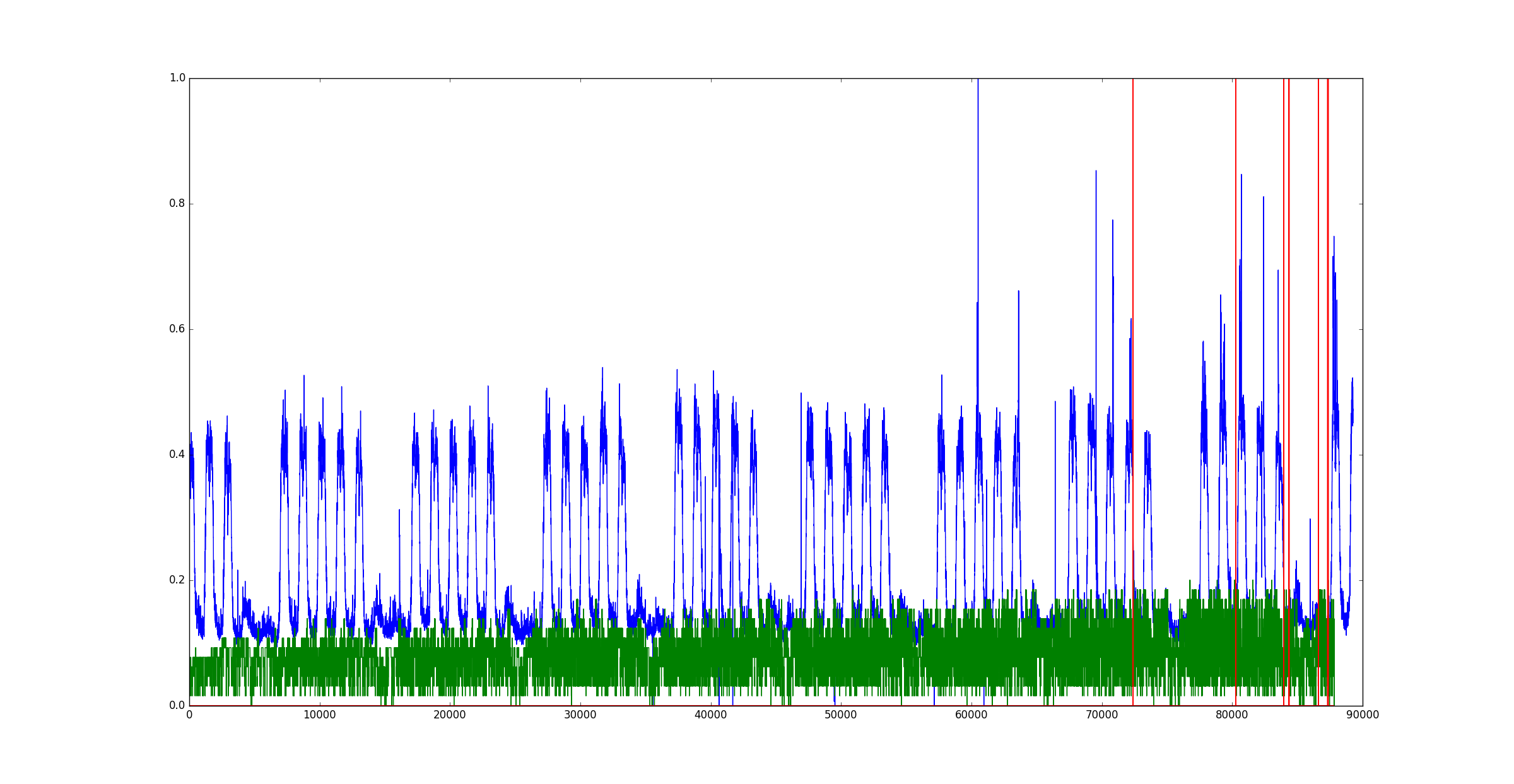}
	\caption{Weekly pattern ignored, anomalies detected when average value runs too high (Sequitur)}
\end{figure}

While fast and capable of catching most anomalies, as well as immune to cyclic phenomena, Sequitur relies on the low compressibility of anomalies to spot them. Therefore, it completely fails to detect any anomaly that results in a \textit{simplified} pattern, like the one in figure \ref{cpu2}. An interesting feature is that the running time of the algorithm depends more on the complexity (i.e. the number of rules generated by Sequitur) than the number of data points. We found it to be, performance-wise and in terms of detection speed, the best algorithm, requiring little to no look-ahead (in contrast to Chaos Game) and insensitive to cyclic variations (compare correct detection in figure \ref{seqok} with errors in figure \ref{chaosfail}). However, the failure mode we observed is not compatible with a real production environment, in which it can indicate a serious failure or an attack.

Therefore, if Sequitur is to be used, it must be coupled with another algorithm able to detect such failures. It is worth noting that these stationary patterns are only dangerous when they happen at \textit{high} values, e.g. 100\% CPU. Thus a simple threshold-based failure detector would complement Sequitur in a satisfactory way.

\subsection{Chaos Game}

\begin{figure}
	\label{chaosfail}
	\centering
	\includegraphics[width=15cm, height=4.5cm]{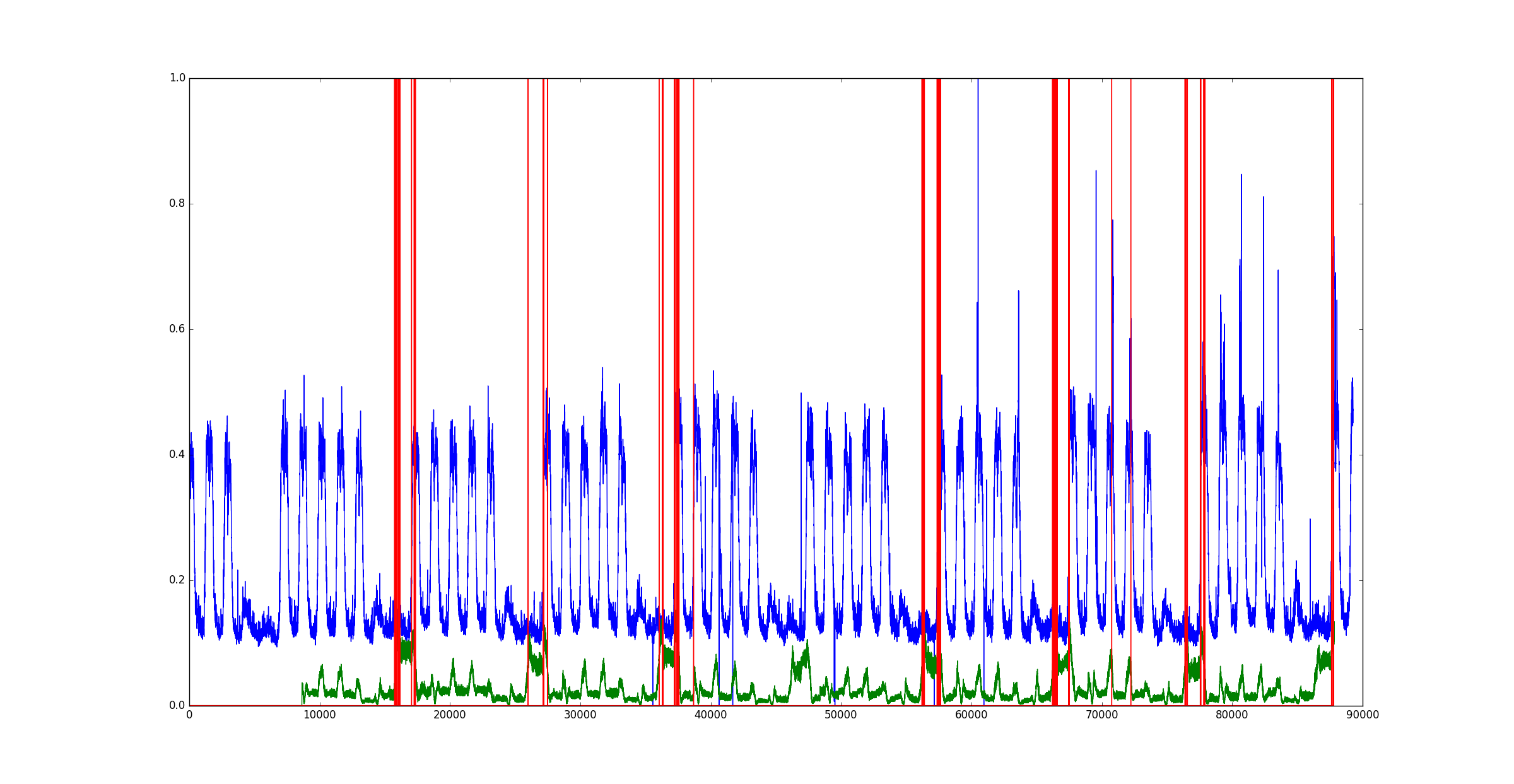}
	\caption{Wrongly detected anomalies due to a weekly cycling pattern (Chaos Game)}
\end{figure}

The Chaos Game algorithm is the most precise anomaly detector with acceptable runtime performance. We found no failure mode in our dataset. The running time is strictly proportional to the length of the time series, and therefore predictable. We found, however, that the look-ahead required to correctly perform anomaly detection (at least twice the feature window) is unacceptably long, and a short feature window makes the algorithm very sensitive to cyclic patterns. See figure \ref{chaosfail} for an example of false positive (correctly handled by Sequitur in figure \ref{seqok}).

We could not find a correct tuning capable of ignoring these low-frequency variations without lengthening the window (and hence the time between an anomaly happening and it being detected) to at least a day. The only way to use the Chaos Game algorithm would be to couple it with a filter capable of rejecting the false positive after detection, or to remove the low-frequency components before analysis, which could potentially lead to a high false negative rate.

\chapter{Conclusion and perspectives}

After implementing and comparing various anomaly detection algorithms based on the SAX representation, we found that none is a perfect fit for real-time detection. This is due partly to the representation itself, which needs time to take a new point into account (due to its heavy downsampling and sliding window analysis). However, tuning the parameters for a good detection led to the conclusion that the feature window is bound to be too large for a practical analysis: for typical weekly patterns found in network monitoring, a good feature window is about one day long. This in turn leads to huge SAX words, with a single symbol representing an hour of data. We conclude that, while extremely useful for fast, ``post-mortem'' analysis, algorithms based on SAX are mostly unsuitable for the kind of real-time analysis we want to perform.

In light of these findings, we did not perform a benchmark of pattern mining algorithms, as the same limitations are bound to apply to them as well. However, given the low computational cost and acceptable results in our anomaly detection experiments, we suggest that hybrid algorithms, working on long time scales with symbolic representation and in real-time with the raw data, could be created to address the overly long detection delay. Conversely, hybrid representations containing both aggregate and raw (or quantized) data could improve detection times.

\chapter{Scientific validation}
\label{chap:scientific_val}

This work has been reviewed and accepted for publication as a technical paper of the Complex System-Digital Campus by:
\begin{itemize}
	\item Ismaila Diouf, McF, Universit\'e Cheikh Antar Diop de Dakar (S\'en\'egal)
	\item Dominique Pastor, Professeur des Universit\'es, T\'el\'ecom Bretagne (France)
\end{itemize}

\printbibliography

\newpage    

\printpublisher

\end{document}